\documentclass[10pt,twocolumn,letterpaper]{article}

\usepackage[pagenumbers]{cvpr}   

\usepackage[dvipsnames]{xcolor}

\newcommand\nnfootnote[1]{%
  \begin{NoHyper}
  \renewcommand\thefootnote{}\footnote{#1}%
  \addtocounter{footnote}{-1}%
  \end{NoHyper}
}

\usepackage{mathtools}

\DeclareMathOperator*{\argmin}{arg\,min}

\definecolor{cvprblue}{rgb}{0.21,0.49,0.74}
\usepackage[pagebackref,breaklinks,colorlinks,citecolor=cvprblue]{hyperref}

\def\paperID{47} 
\def\confName{CVPR}
\def\confYear{2024}

\title{Do Counterfactual Examples Complicate Adversarial Training?}

\author{
Eric Yeats$^1$ \quad Cameron Darwin \quad Eduardo Ortega$^2$ \quad Frank Liu$^3$ \quad Hai Li$^1$ \\
$^1$Duke University \quad $^2$Arizona State University \quad $^3$Old Dominion University \\
}

\begin{document}
\maketitle
\begin{abstract}
We leverage diffusion models to study the robustness-performance tradeoff of robust classifiers. Our approach introduces a simple, pretrained diffusion method to generate low-norm counterfactual examples (CEs): semantically altered data which results in different true class membership. We report that the confidence and accuracy of robust models on their clean training data are associated with the proximity of the data to their CEs. Moreover, robust models perform very poorly when evaluated on the CEs directly, as they become increasingly invariant to the low-norm, semantic changes brought by CEs. The results indicate a significant overlap between non-robust and semantic features, countering the common assumption that non-robust features are not interpretable.
\end{abstract}

\begin{figure}[t!]
    \centering
    \includegraphics[width=\columnwidth]{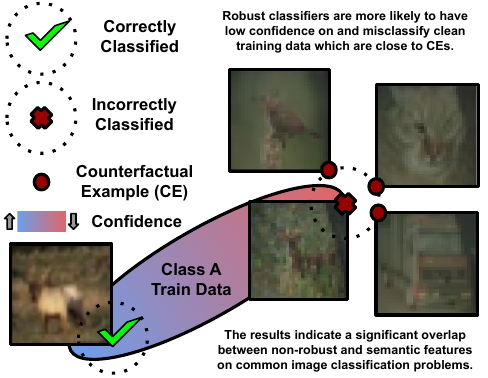}
    \caption{Conceptual depiction of the relationship between accuracy of robustly trained models with proximity to counterfactual examples (CEs). Stronger adversarial training inevitably leads to misclassification of some clean training data, incurring downstream test performance loss. We hypothesize that adversarially trained models are forced to become invariant to some semantic features due to the nearby presence of true CEs.}\label{fig:ces_manifolds}
\end{figure}

\section{Introduction}


Leading theory by \citet{ilyas2019adversarial} asserts that adversarial vulnerability arises from reliance of DNN classifiers on \textit{non-robust} features: well-generalizing, yet brittle features which are not comprehensible to humans. As robust models must become invariant to this subset of predictive features, they inevitably lose performance \cite{tsipras2018robustness}. However, humans appear to break this trend by simultaneously maintaining good performance and robustness on adversarial data \cite{ross2018improving}.

The existence of adversarial perturbations reflects the stark differences between the perception of DNNs and our own. DNNs are notoriously uninterpretable \cite{rudin2019stop, yeats2022nashae}, and there is a need for interpretable decision-making in high-stakes scenarios. One model-agnostic way to interpret classifier decision-making is to generate \textit{counterfactual examples} (CEs): subtle, semantic changes to an input datum which would result in a classifier predicting a target class \cite{verma2020counterfactual}. Following their definition from psychology, CEs for an image-label pair $(x, y)$ can be described by the statement ``if $y'$ were the true class of image $x$ (instead of $y$), then $x$ would look like $x'$.'' While CEs can help users understand the decision-making of DNN classifiers, practical methods to generate CEs are problematically similar to those for adversarial attacks \cite{freiesleben2022intriguing, pawelczyk2022exploring}. Hence, methods to generate CEs for DNNs often require robust models in some form \cite{boreiko2022sparse, augustin2022diffusion}.

We develop new understanding of the robustness-performance tradeoff through our study of the semantic feature distributions learned by robust classifiers. Our study leverages independently generated CE datasets which we create with a simple, pretrained diffusion model technique. We report that robust models, unlike standard (non-robust) models, are more likely to lose confidence on and misclassify \textit{clean} training data which have nearby CEs, and robust models become invariant to the semantic changes brought by CEs. Contrary to common assumptions, our findings suggest significant overlap between \textit{non-robust} and \textit{semantically meaningful} features. This conflict speaks to the complexity of the adversarial problem and motivates alternative approaches to robustness which can resolve or avert it.\nnfootnote{Corresponding Author Email: {\tt eric.yeats@duke.edu}}

\section{Background}

\paragraph{Adversarial Examples and Adversarial Training}

Adversarial examples are minute, malicious data perturbations which alter classifier predictions \cite{szegedy2013intriguing}. For a classifier $f_\theta:\mathcal{X}\rightarrow\mathcal{Y}$ where $\mathcal{X}\in\mathbb{R}^n$ and $\mathcal{Y}$ is the space of categorical distributions of support cardinality $k$, $\tilde{y}$-targeted adversarial examples $\tilde{x}$ are often computed by altering an input $x$ to descend the negative log loss \cite{goodfellow2014explaining} $\mathcal{L}: \Theta \times \mathcal{X} \times \mathcal{Y} \rightarrow \mathbb{R}^+$:
\begin{equation}
    \tilde{x} = \argmin_{x' \in \mathcal{B}_\varepsilon(x)} \mathcal{L}(\theta, x', \tilde{y}),
\end{equation}
where the neighborhood $\mathcal{B}_\varepsilon(x)$ is commonly defined with the $L^\infty$ or $L^2$ norm. In this work, we focus on $L^2$ adversarial examples. Adversarial training consists of finding the optimal model parameters $\theta^*$ for the robust objective \cite{madry2017towards}:
\begin{equation}
    \theta^* = \argmin_\theta\mathbb{E}_{x,y}\max_{x' \in \mathcal{B}_\varepsilon(x)} \mathcal{L}(\theta, x', y).
\end{equation}
There are other robustness methods such as gradient regularization \cite{yeats2021improving, finlay2021scaleable} and randomized smoothing \cite{cohen2019certified}, but adversarial training is the most well-known and successful.

\paragraph{Diffusion Models and Classifier-free Guidance} Diffusion models learn a sequence of functions $\epsilon_\theta(x_t, t)$ which predict the noise added to data by a diffusion process at time $t \in [0, 1]$ \cite{ho2020denoising}. To generate new data, $\epsilon_\theta(x_t, t)$ is used to iteratively denoise samples $x_t$ initially drawn from a known noise prior. The noise prediction function $\epsilon_\theta(x_t, t)$ is known to approximate the negative score at diffusion step $t$ \cite{song2020score,yeats2023adversarial} (i.e., $\epsilon_\theta(x_t, t) \approx -\sigma_t\nabla_x \log p_t(x_t)$). For class-conditional diffusion models, this property can be used to reproduce the effect of adding targeted classifier gradients \cite{ho2022classifier}:
\begin{multline}
    \epsilon^w_\theta(x_t, y, t) \coloneq (w+1)\epsilon_\theta(x_t, y, t) - w \epsilon_\theta(x_t, t) \\ \approx \epsilon_\theta(x_t, y, t) + w \sigma_t\nabla_x \mathcal{L}(\theta_t, x_t, y),
\end{multline}\label{eqn:cls_free_diff}
where $\theta_t$ are the equivalent parameters for a time-conditional classifier. Coined ``classifier-free guidance'' with weight $w$, sampling with $\epsilon^w_\theta(x_t, y, t)$ as the noise prediction function generates high-quality, conditional data.

\section{Methodology}

\paragraph{CE Dataset Generation}

Our approach to CE generation for datum $x$ is to recast it as sampling from a sequence of un-normalized distributions defined by the product of the data distribution (represented by the class-conditional diffusion model) with an independently diffused CE ``neighborhood'' distribution of scale $\sigma_{CE}$ centered on $\mu_{CE}=x$. The score of this distribution at each time $t$ is the sum of scores of the two components, so we may simply add the diffused neighborhood score to the diffusion model score.

We present two variants of the noise prediction function corresponding to different choices for the neighborhood distribution: Gaussian and Boltzmann-inspired (see appendix).
\begin{equation}
    \epsilon^G_\theta(x_t, y, \mu_{CE}, t) \coloneq \epsilon^w_\theta(x_t, y, t) - \frac{\alpha_t \mu_{CE} - x_t}{\alpha_t^2 \sigma_{CE}^2 + \sigma_t^2}
\end{equation}
\begin{multline}
    \epsilon^B_\theta(x_t, y, \mu_{CE}, t) \coloneq \\
    \epsilon^w_\theta(x_t, y, t) -\frac{\sqrt{2}}{\alpha_t \sigma_{CE}} \text{hardtanh}(\gamma_t\,(x_t - \alpha_t \mu_{CE}))
\end{multline}
where $\alpha_t$ is a (diffusion) time-dependent scalar decreasing in $t \in [0, 1]$, $\sigma_t = \sqrt{1 - \alpha_t^2}$, and $\gamma_t$ is a time-dependent scalar derived from the first-order Maclaurin series of the Boltzmann-inspired scores, and is defined as:
\begin{equation}
    \gamma_t \coloneq \frac{\sqrt{2}}{\alpha_t \sigma_{CE}} - \frac{\sqrt{2}}{\sigma_t \sqrt{\pi}}\left(\exp{\left(\frac{\sigma_t^2}{\alpha_t^2\sigma^2_{CE}}\right)}\,\text{erfc}\left(\frac{\sigma_t}{\alpha_t\sigma_{CE}}\right)\right)^{-1}.
\end{equation}
A CE $x_{CE}$ is generated for a datum $x$ by assigning $\mu_{CE}\leftarrow x$ and sampling with $\epsilon^G_\theta(\cdot)$ or $\epsilon^B_\theta(\cdot)$ with guidance towards target class $y_{CE}$. Guidance $w$ and neighborhood scale $\sigma_{CE}$ are hyperparameters. The Boltzmann-inspired distribution has a sharper mode than the Gaussian distribution, encouraging $x-x_{CE}$ to be lower norm ($L^2$) and more sparse.

\begin{figure}[tb]
    \centering
    \hfill
    \begin{subfigure}[b]{0.48\columnwidth}
        \includegraphics[width=\textwidth]{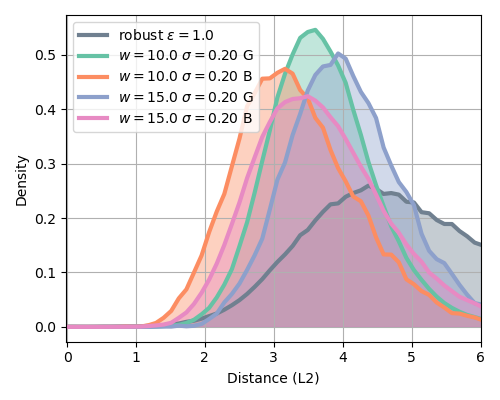}
        \caption{CIFAR10 CE Dist. ($L^2$)}\label{fig:cifar_ce_dc_dists}
    \end{subfigure}
    \hfill
    \begin{subfigure}[b]{0.48\columnwidth}
        \includegraphics[width=\textwidth]{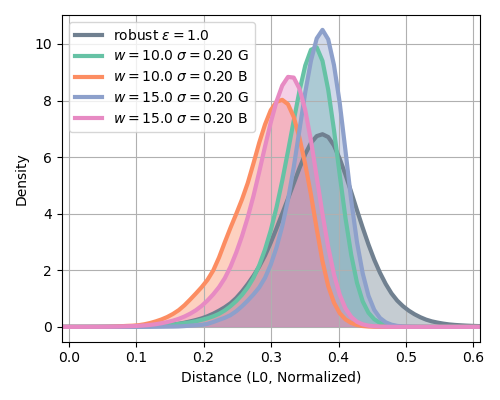}
        \caption{CIFAR10 CE Dist. ($L^0$)}\label{fig:cifar_ce_dc_l0_dists}
    \end{subfigure}
    \hfill

    \hfill
    \begin{subfigure}[b]{0.48\columnwidth}
        \includegraphics[width=\textwidth]{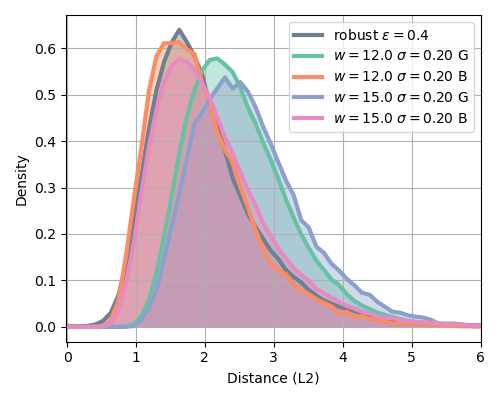}
        \caption{SVHN CE Dist. ($L^2$)}\label{fig:svhn_ce_dc_dists}
    \end{subfigure}
    \hfill
    \begin{subfigure}[b]{0.48\columnwidth}
        \includegraphics[width=\textwidth]{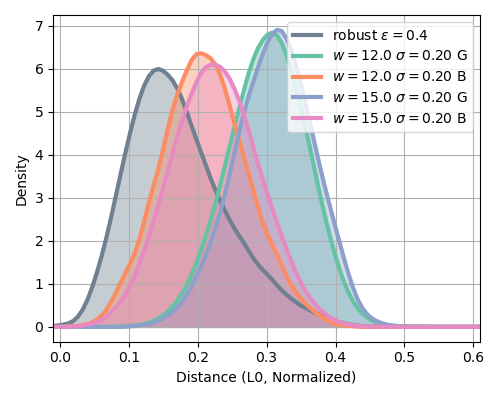}
        \caption{SVHN CE Dist. ($L^0$)}\label{fig:svhn_ce_dc_l0_dists}
    \end{subfigure}
    \hfill
    \caption{CE distribution comparison. Boltzmann variant CEs produce lower-norm, sparser changes. Best viewed in color.}\label{fig:ce_comparison}
\end{figure}

\section{Experiments}

\begin{figure*}[ht!]
    \centering
    \hfill
    \begin{subfigure}[b]{0.32\textwidth}
        \includegraphics[width=\textwidth, trim={0 0 0 25}, clip]{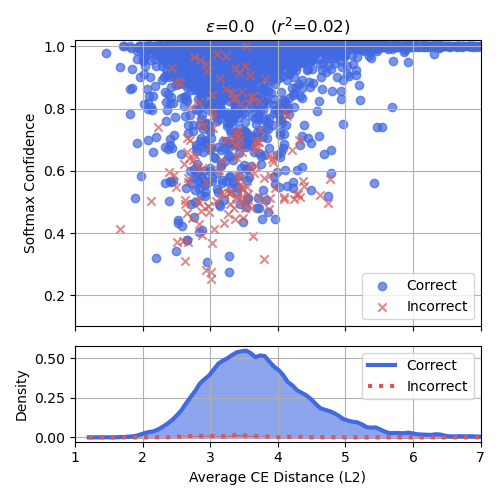}
        \caption{CIFAR10 $\varepsilon=0$\\$r^2=0.02$}\label{fig:ce_scatter_cifar_a}
    \end{subfigure}
    \hfill
    \begin{subfigure}[b]{0.32\textwidth}
        \includegraphics[width=\textwidth, trim={0 0 0 25}, clip]{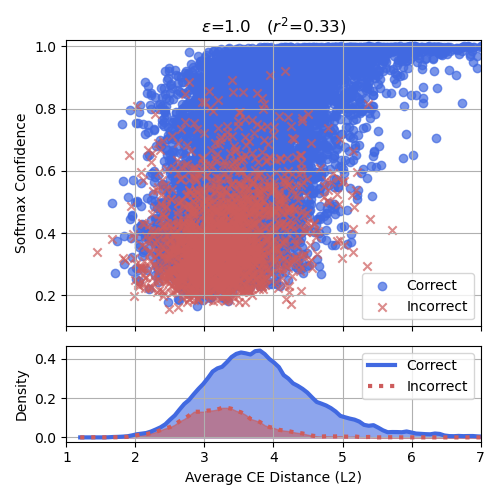}
        \caption{CIFAR10 $\varepsilon=1$\\$r^2=0.33$}\label{fig:ce_scatter_cifar_b}
    \end{subfigure}
    \hfill
    \begin{subfigure}[b]{0.32\textwidth}
        \includegraphics[width=\textwidth, trim={0 0 0 25}, clip]{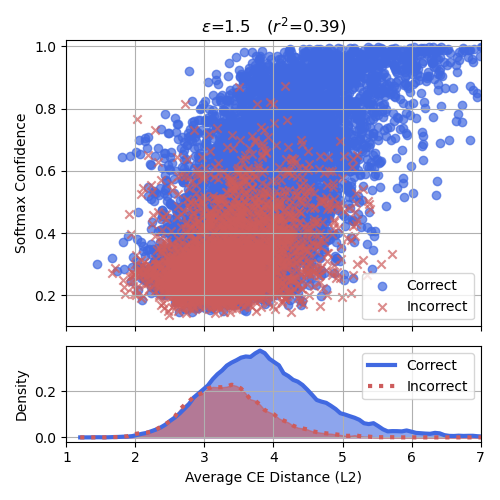}
        \caption{CIFAR10 $\varepsilon=1.5$\\$r^2=0.39$}\label{fig:ce_scatter_cifar_c}
    \end{subfigure}
    \hfill

    \hfill
    \begin{subfigure}[b]{0.32\textwidth}
        \includegraphics[width=\textwidth, trim={0 0 0 25}, clip]{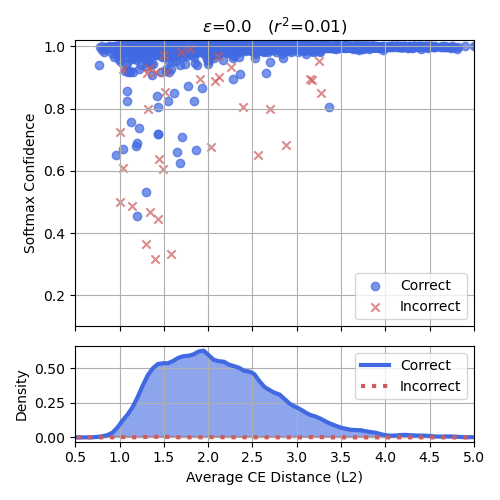}
        \caption{SVHN $\varepsilon=0$\\$r^2=0.01$}\label{fig:ce_scatter_svhn_a}
    \end{subfigure}
    \hfill
    \begin{subfigure}[b]{0.32\textwidth}
        \includegraphics[width=\textwidth, trim={0 0 0 25}, clip]{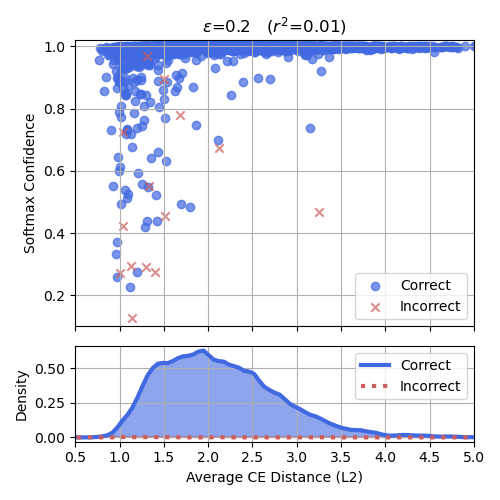}
        \caption{SVHN $\varepsilon=0.2$\\$r^2=0.01$}\label{fig:ce_scatter_svhn_b}
    \end{subfigure}
    \hfill
    \begin{subfigure}[b]{0.32\textwidth}
        \includegraphics[width=\textwidth, trim={0 0 0 25}, clip]{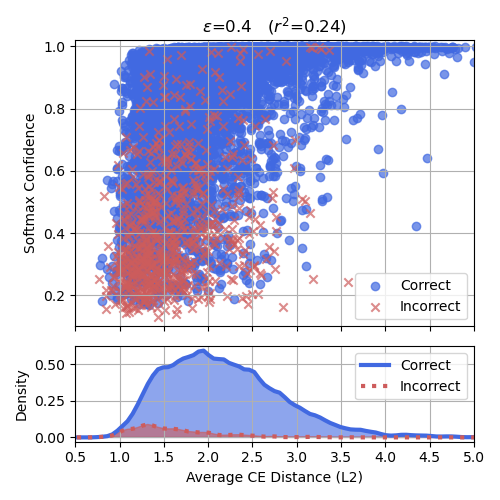}
        \caption{SVHN $\varepsilon=0.4$\\$r^2=0.24$}\label{fig:ce_scatter_svhn_c}
    \end{subfigure}
    \hfill

    \caption{Scatter plots of classifier confidence and average CE distance of 10000 clean training samples as adversarial training norm is increased. Robust models are more likely to misclassify and lose confidence on data which have closer CEs. Best viewed in color.}\label{fig:ce_scatter}
\end{figure*}

\begin{figure*}[tb]
    \centering
    \hfill
    \begin{subfigure}[b]{0.59\textwidth}
        \includegraphics[width=\textwidth]{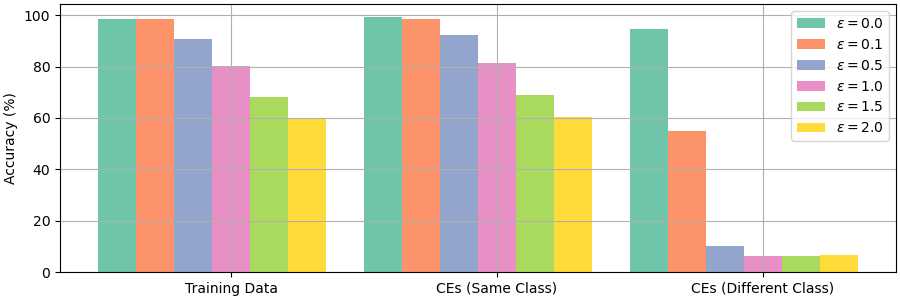}
        \caption{CIFAR10 Accuracy}\label{fig:ce_acc_bar_cifar}
    \end{subfigure}
    \hfill
    \begin{subfigure}[b]{0.39\textwidth}
        \includegraphics[width=\textwidth]{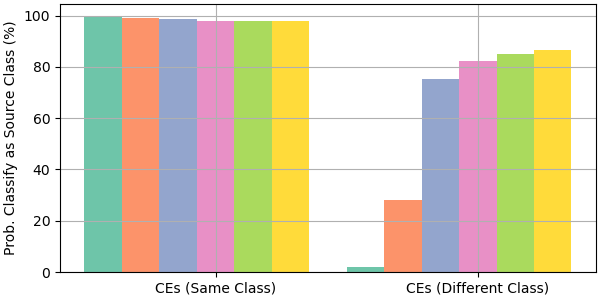}
        \caption{CIFAR10 Prob. Source Predicted}\label{fig:ce_sc_prob_bar_cifar}
    \end{subfigure}
    \hfill

    \hfill
    \begin{subfigure}[b]{0.59\textwidth}
        \includegraphics[width=\textwidth]{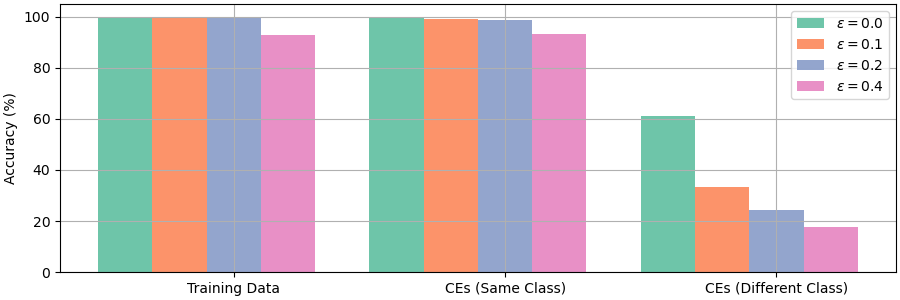}
        \caption{SVHN Accuracy}\label{fig:ce_acc_bar_svhn}
    \end{subfigure}
    \hfill
    \begin{subfigure}[b]{0.39\textwidth}
        \includegraphics[width=\textwidth]{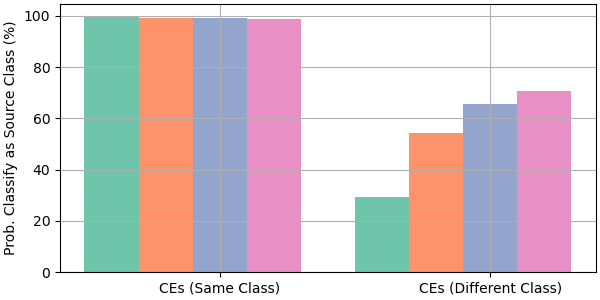}
        \caption{SVHN Prob. Source Predicted}\label{fig:ce_sc_prob_bar_svhn}
    \end{subfigure}
    \hfill
    \caption{Classifier performance on 10000 training data and 200000 CE data generated from the training samples. Best viewed in color.}\label{fig:ce_acc}
\end{figure*}

We run experiments in the PyTorch framework \cite{paszke2019pytorch} with the CIFAR10 \cite{krizhevsky2009learning} and SVHN \cite{netzer2011reading} image classification benchmarks . We employ conditional score-based models (SBMs) \cite{song2020score} for all diffusion models and $L^2$ PGD($8$) \cite{madry2017towards} $2\times$WideResNet-40 models \cite{zagoruyko2016wide} for all robust classifiers.
\paragraph{CE Dataset Evaluation} In each experiment we generate 2 CEs for each class for at least 1000 samples from the training set, resulting in at least 20000 CEs. We compare the two CE generation variants with CEs generated by a robust classifier in \cref{fig:ce_comparison}. The Boltzmann variant (type $B$) produces CEs with lower-norm ($L^2$) and sparser ($L^0$) changes than the Gaussian variant does. The Boltzmann variant produces CEs of norm less than or equal to the norm of CEs produced by the robust classifier. Standard classifiers (no adversarial training) achieve high accuracy on the CEs (i.e., they predict $y_{CE}$ when provided $x_{CE}$), indicating good semantic quality. The remaining experiments use 200000 Boltzmann variant CEs generated from 10000 training samples (CIFAR10 or SVHN) with $w=15$ and $\sigma_{CE}=0.2$. Please see the appendix for more information.
\paragraph{Robust Model Evaluation} Our experience is that performance loss of robust models begins on the clean \textit{training} data, incurring performance loss on clean test data downstream. We plot the average $L^2$ distance of clean training samples with the confidence and accuracy of robust models on the clean samples in \cref{fig:ce_scatter}. Standard models ($\varepsilon=0$) achieve very high clean training accuracy, as expected. As robust training budget $\varepsilon$ increases, robust models are more likely to misclassify clean training data which are closer to their CEs. Moreover, the confidence of robust models on their training data becomes correlated with the proximity of the data to their CEs ($r^2=0.39$ for CIFAR10 and $r^2=0.24$ for SVHN).

\cref{fig:ce_acc_bar_cifar} and \cref{fig:ce_acc_bar_svhn} depict the accuracy of robust models on clean train data and on CEs generated from the train data. Same-class CEs resemble the original data; hence their robust accuracy trends are similar. However, robust models perform very poorly on different-class CEs. \cref{fig:ce_sc_prob_bar_cifar} and \cref{fig:ce_sc_prob_bar_svhn} depict the probability that the original label $y$ is predicted by a robust model for $(x_{CE}, y_{CE})$, given that the robust model was correct on $(x, y)$. Robust models are much more likely to misclassify CEs as having the source label, indicating that they become invariant to the low-norm, semantic changes brought by CEs.

\begin{figure}[t!]
    \centering
    \includegraphics[width=0.9\columnwidth]{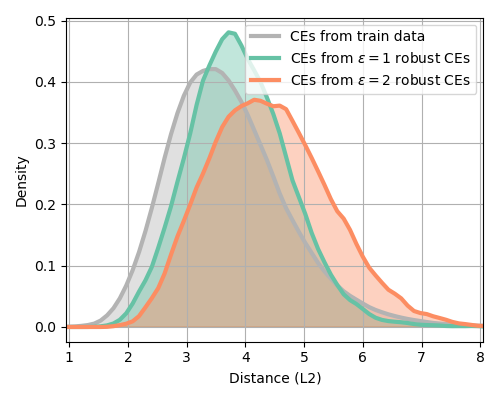}
    \caption{Distance of different-class CEs generated by the Boltzmann method ($w=15\ \sigma_{CE}=0.2$) when the input data is the original CIFAR10 train data, CEs generated by a robust $\varepsilon=1$ model from the CIFAR10 train data, and CEs generated by a robust $\varepsilon=2$ model from the CIFAR10 train data. Robust model CEs tend to be in data regions farther away from our diffusion-generated CEs. Best viewed in color.}\label{fig:ce_from_ce}
\end{figure}

\cref{fig:ce_from_ce} depicts the $L^2$ distance distribution of Boltzmann-type CEs generated from the original CIFAR10 training data and Boltzmann-type CEs generated from robust CEs ($\varepsilon=1$, $\varepsilon=2$). On average, Boltzmann-type CEs are farther away from their source data samples when the data samples are robust model CEs (compared to using the original training data as the source data). Since robust model CEs are generated by following input gradient to maximize the confidence of a target class, this indicates that robust model gradients orient towards data regions which are farther away from CEs. Future work may investigate a link between the perceptually aligned gradient of robust classifiers \cite{tsipras2018robustness} and the proximity of data to CEs.

The method presented in this work can be leveraged to convert a conditional diffusion model into an interpretable classifier. Using the diffusion CE method with a ``class with lowest average CE distance'' decision rule achieves $\sim 85\%$ accuracy on $1000$ input samples of  the CIFAR10 test set. Critically, input data for classification are never provided directly to the diffusion model - the input data merely guides CE sampling with analytic scores. 

\paragraph{Limitations} Our study considers one combination of classifier architecture, diffusion model, and adversarial attack. This study is limited to the $L^2$ norm constraint and two image classification benchmarks. Investigation with additional datasets, attacks, and models is left to future work.

\section{Conclusion}

We present a simple, classifier-free diffusion method to generate counterfactual examples (CEs) which enables novel investigation of the performance loss of robust classifiers. Our results indicate a significant overlap between \textit{non-robust} and \textit{semantically meaningful} features, countering the common assumption that \textit{non-robust} features are not interpretable. Hence, robust models must become invariant to this subset of \textit{semantic} features along with \textit{non-semantic} adversarial perturbations. Our study motivates new approaches to robust training which can resolve this issue.

\section*{Acknowledgements}

Part of this research was conducted while Eric Yeats was supported by a DOE-SCGSR Award at Oak Ridge National Laboratory (ORNL). Hence, this material is based upon work supported by the U.S. Department of Energy, Office of Science, Office of Workforce Development for Teachers and Scientists, Office of Science Graduate Student Research (SCGSR) program. The SCGSR program is administered by the Oak Ridge Institute for Science and Education (ORISE) for the DOE. ORISE is managed by ORAU under contract number DE-SC0014664. All opinions expressed in this paper are the author’s and do not necessarily reflect the policies and views of DOE, ORAU, or ORISE. This research was also supported in part by the U.S. Department of Energy, through the Office of Advanced Scientific Computing Research's “Data-Driven Decision Control for Complex Systems (DnC2S)” project.

This research is supported in part by the U.S. National Science Foundation funding CNS-1822085 and U.S. Air Force Research Lab funding FA8750-21-1-1015.

This research used resources of the Experimental Computing Laboratory (ExCL) at the Oak Ridge National Laboratory, which is supported by the Office of Science of the U.S. Department of Energy under Contract No. DE-AC05-00OR22725.

{
    \small
    \bibliographystyle{ieeenat_fullname}
    \bibliography{main}
}

\clearpage
\setcounter{page}{1}
\maketitlesupplementary\

\section{Reproducibility Information}

\begin{figure}[tb]
    \centering
    \hfill
    \begin{subfigure}[b]{0.48\columnwidth}
        \includegraphics[width=\textwidth]{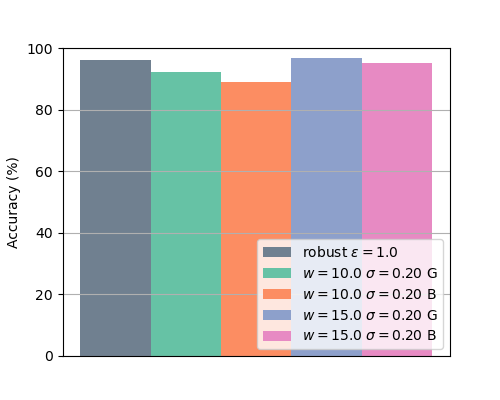}
        \caption{CIFAR10 Standard Accuracy}\label{fig:cifar_ce_acc}
    \end{subfigure}
    \hfill
    \begin{subfigure}[b]{0.48\columnwidth}
        \includegraphics[width=\textwidth]{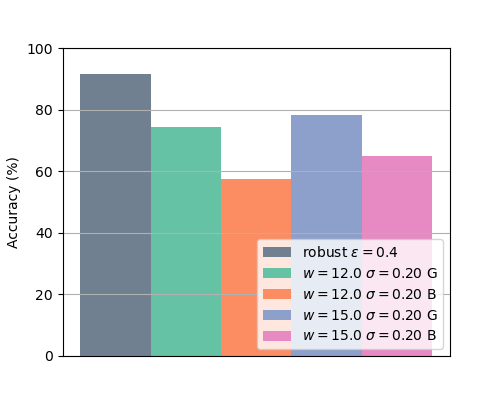}
        \caption{SVHN Standard Accuracy}\label{fig:svhn_ce_acc}
    \end{subfigure}
    \hfill
    \caption{Comparison of the accuracy attained by standard models on the CEs. Best viewed in color.}\label{fig:ce_acc_comparison}
\end{figure}

$L^2$ distance for the CEs is defined as $\|x-x_{CE}\|_2$, where $x$ is the original sample from which the CE $x_{CE}$ was generated. $L^0$ distances ($\|x-x_{CE}\|_0$) employ an element-wise threshold of $0.02$ for each pixel difference value, and the reported value is normalized (the thresholded $L^0$ distance is divided by the dimension of $x$). CE quality is measured by the accuracy of a standard classifier in predicting $y_{CE}$ given $x_{CE}$ (depicted in \cref{fig:ce_acc_comparison}). On SVHN there appears to be sizeable drop from robust CE accuracy to the Boltzmann CEs. This may be due to several reasons. First, the $\epsilon=0.4$ robust classifier may be making adversarial changes (rather than semantic), biasing its quality measure higher at that $L^2$ distance. Second, diffusion models tend to ignore class conditioning information when their inputs are simple (like those of SVHN), and they operate instead as unconditional denoising functions. This would cause the CE generation method to fail. Visual inspection of the CEs for SVHN indicates that the $\varepsilon=0.4$ classifier is making some adversarial (non-semantic) changes, and the diffusion method fails in some cases. Please refer to the end of the supplementary material for visual depictions of CEs.

\subsection*{Robust WideResNet Experiments}

2$\times$WideResNet-40 models \cite{zagoruyko2016wide} are adversarially trained with PGD(8) \cite{madry2017towards} in $L^2$ on CIFAR10 \cite{krizhevsky2009learning} and SVHN \cite{netzer2011reading} for 100 epochs and learning rate $1e-3$ using the Adam optimizer $\beta=(0.9, 0.999)$. For standard models ($\varepsilon=0$), dropout with rate 0.3 is used. The data augmentations for CIFAR10 are random horizontal flips and random crops with padding 4. No augmentations are used for SVHN. The clean accuracies of the trained models are listed in tables (\ref{tab:cifar_acc}) and (\ref{tab:svhn_acc}). All adversarial attacks are PGD(8).

We observed that robust training on SVHN would collapse for $\varepsilon \geq 0.5$, so only $\varepsilon \leq 0.4$ experiments are reported. Coincidentally, $\varepsilon \sim 0.5$ is half the distance of a large amount of different-class CEs to their original data on SVHN. We hypothesize that the presence of many CEs at the $L^2$ distance $1$ may be related to this training collapse phenomenon, as an adversarial budget of $\varepsilon \sim 0.5$ could make the source class of perturbed training data highly ambiguous.

\begin{table*}[t]
\caption{Accuracy of CIFAR10 Models on the CIFAR10 test set}\label{tab:cifar_acc}
\vskip 0.15in
\begin{center}
\begin{small}
\begin{sc}
\begin{tabular}{lccc}
\toprule
$\varepsilon$ & Clean Accuracy (\%) & \quad\quad\quad & PGD(8, 0.5) Accuracy (\%) \\
\midrule
0 & \textbf{93.08} & & 5.75 \\
0.1 & 91.33 & & 45.06 \\
0.5 & 84.87 & & \textbf{54.73} \\
1.0 & 76.23 & & 54.53 \\
1.5 & 66.21 & & 52.34 \\
2.0 & 58.37 & & 47.53 \\
\bottomrule
\end{tabular}
\end{sc}
\end{small}
\end{center}
\vskip -0.1in
\end{table*}

\begin{table*}[t]
\caption{Accuracy of SVHN Models on the SVHN test set}\label{tab:svhn_acc}
\vskip 0.15in
\begin{center}
\begin{small}
\begin{sc}
\begin{tabular}{lccc}
\toprule
$\varepsilon$ & Clean Accuracy (\%) & \quad\quad\quad & PGD(8, 0.2) Accuracy (\%) \\
\midrule
0 & \textbf{96.75} & & 77.46 \\
0.1 & 95.63 & & 86.42 \\
0.2 & 95.41 & & \textbf{86.84} \\
0.4 & 93.09 & & 85.43 \\
\bottomrule
\end{tabular}
\end{sc}
\end{small}
\end{center}
\vskip -0.1in
\end{table*}

\subsection*{Diffusion Models}

In all experiments using diffusion models, we employ the variance-preserving (VP) score-based models (SBM) of Song et al. \cite{song2020score}. The SBM architecture follows a U-net structure with four symmetric stages of two ResNet blocks in each encoding or decoding stage. Downsampling (and upsampling, respectively) occurs in the innermost three stages (i.e., stages 2, 3, 4). 128 channels (features) are used, and the number of features used is doubled to 256 for stages 2, 3, and 4. Attention is applied at the center of the U-net and after the first downsampling stage and before the last upsampling stage. The SBMs use the SiLU activation function \cite{hendrycks2016gaussian} and GroupNorm \cite{wu2018group} with a group size of 32. Training on CIFAR10 or SVHN occurs for 1 million iterations of batch size 128 with a learning rate of $2e-4$ and Adam optimizer $\beta=(0.9, 0.999)$. We use a learning rate warmup for 5000 iterations and gradient clipping with norm 1. Class conditioning is provided as a learnable embedding which is added to the time condition embedding. An additional learnable null embedding signifies that \textit{no} class information is provided. During conditional SBM training, class conditions are dropped and replaced with this null embedding at a rate of $30\%$.

\section{CE Generation Method}

CEs were generated by sampling from a sequence of un-normalized distributions represented by the product of the data distribution (rep. by diffusion model) and an independently diffused neighborhood distribution. The diffusion model used is a variance-preserving (VP) score-based model of Song et al. \cite{song2020score} and the sampling strategy for generation used an Euler-Maruyama predictor with 1000 discrete steps. 

Since the density at each step in the sequence is represented as the product of the diffused data distribution and an independently diffused neighborhood distribution, sampling amounts to adding the conditional diffusion score (and classifier free guidance) with the analytic neighborhood diffusion score. Sampling then proceeds as normal (see \cite{song2020score}) with the augmented score.

\paragraph{Robust Model CEs} For CE generation with robust models, we push data in the direction of targeted gradients with a step size of $0.05$ until a targeted class confidence of $0.9$ or a maximum of 200 steps is reached. At each step, we clip the pixels of the image to remain in [0, 1].

\subsection*{Boltzmann-Inspired Distribution}

The density of the 1D Boltzmann-inspired distribution proposed in this work is given by:

\begin{equation}
    b(x) = \frac{1}{\sqrt{2}\,\sigma_{CE}}\exp{\left(-\frac{\sqrt{2}\,|x-\mu_{CE}|}{\sigma_{CE}}\right)},
\end{equation}

where $\mu_{CE}$ is its mean and $\sigma_{CE}$ is its standard deviation. Scaling samples of this distribution with $\alpha_t > 0$ amounts to sampling from a new distribution with mean $\alpha_t\mu_{CE}$ and standard deviation $\alpha_t\sigma_{CE}$. Defining $y_t=x-\alpha_t\mu_{CE}$, we convolve this distribution with a Gaussian distribution of mean $\mu=0$ and scale $\sigma_t$ to yield the (diffusion) time-dependent distribution:

\begin{multline}
    b_t(x) = \\
    \frac{\exp\left(\frac{\sigma_t^2}{\alpha_t^2\sigma_{CE}^2} - \frac{y_t}{\sqrt{2}\alpha_t\sigma_{CE}}\right) \text{erfc}\left(\frac{\sigma_t}{\alpha_t\sigma_{CE}} - \frac{y}{\sqrt{2}\sigma_t}\right)}{2 \sqrt{2} \alpha_t\sigma_{CE}} \\
    + \frac{\exp\left(\frac{\sigma_t^2}{\alpha_t^2\sigma_{CE}^2} + \frac{y_t}{\sqrt{2}\alpha_t\sigma_{CE}}\right) \text{erfc}\left(\frac{\sigma_t}{\alpha_t\sigma_{CE}} + \frac{y}{\sqrt{2}\sigma_t}\right)}{2 \sqrt{2} \alpha_t\sigma_{CE}}.
\end{multline}\label{eqn:diffused_boltz}

Taking the logarithm of $b_t(x)$ and differentiating with respect to $x$, we have:

\begin{multline}
    \nabla_x \log b_t(x) = \\
    \frac{\sqrt{2}}{\alpha_t\sigma_{CE}}\Bigg(e^{\frac{\sqrt{2}}{\alpha_t\sigma_{CE}}y_t}\text{erfc}\left(\frac{\sigma_t}{\alpha_t\sigma_{CE}} + \frac{y_t}{\sqrt{2}\sigma_t}\right) \\
    - e^{-\frac{\sqrt{2}}{\alpha_t\sigma_{CE}}y_t}\text{erfc}\left(\frac{\sigma_t}{\alpha_t\sigma_{CE}} - \frac{y_t}{\sqrt{2}\sigma_t}\right) \Bigg) \Bigg/ \\
    \Bigg( e^{\frac{\sqrt{2}}{\alpha_t\sigma_{CE}}y_t}\text{erfc}\left(\frac{\sigma_t}{\alpha_t\sigma_{CE}} + \frac{y_t}{\sqrt{2}\sigma_t}\right) \\
    + e^{-\frac{\sqrt{2}}{\alpha_t\sigma_{CE}}y_t}\text{erfc}\left(\frac{\sigma_t}{\alpha_t\sigma_{CE}} - \frac{y_t}{\sqrt{2}\sigma_t}\right)\Bigg).
\end{multline}

Although this expression yields the exact scores for the diffused Boltzmann-inspired distribution in 1D, it is numerically unstable. We note that $\nabla_x \log b_t(x)$ is sigmoidal, and we elect to approximate it using the hardtanh function. One may consider the $\frac{\sqrt{2}}{\alpha_t \sigma_{CE}}$ expression to define the range of the score values, and the remainder of the expression defines a Gaussian-like score near the mean. Hence, we define our approximate scores as:

\begin{equation}
    \nabla_{x} \log b_t(x) \approx \frac{\sqrt{2}}{\alpha_t \sigma_{CE}} \text{hardtanh}(\gamma_t\,y_t),
\end{equation}

where $\gamma_t$ is the first term of a Maclaurin series estimate of $\nabla_x \log b_t(x)$ and is given by:

\begin{equation}
    \gamma_t = \frac{\sqrt{2}}{\alpha_t \sigma_{CE}} - \frac{\sqrt{2}}{\sigma_t \sqrt{\pi}}\left(\exp{\left(\frac{\sigma_t^2}{\alpha_t^2\sigma^2_{CE}}\right)}\,\text{erfc}\left(\frac{\sigma_t}{\alpha_t\sigma_{CE}}\right)\right)^{-1}.
\end{equation}

With $u=\frac{\sigma_t}{\alpha_t\sigma_{CE}}$, the expression $\frac{\exp(-u^2)}{\text{erfc}(u)}$ is numerically unstable for large values of $u$. However, beyond a certain point (e.g., $u \geq 20$), the function behaves as a linear function with slope $\sqrt{\pi}$. We avoid the numerical instability by switching to a linear approximation at $u=20$.

\begin{figure}[b]
    \centering
    \includegraphics[width=0.99\columnwidth]{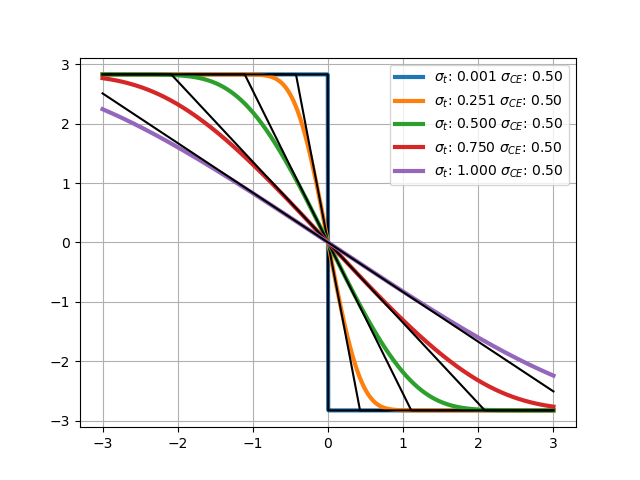}
    \caption{Comparison of true 1D Boltzmann-inspired scores with the proposed hardtanh approximation (black). The mean $\mu_{CE}$ is selected to be 0.  Best viewed in color.}\label{fig:boltz_score_approx}
\end{figure}

The exact score function $\nabla_x \log b_t(x)$ and its approximation is displayed with $\alpha_t=1$ for various values of $\sigma_t$ and in figure \ref{fig:boltz_score_approx}. In our experiments, this 1D score function is applied element-wise to vector inputs, providing a sparsifying effect on CE generation.

For completeness, we include $\nabla^2_x \log b_t(x)$, which was evaluated at $y_t=0$ to yield the first term of the Maclaurin series of $\nabla_x \log b_t(x)$:

\begin{multline}
    \nabla^2_x \log b_t(x) = \\
    \Bigg[\frac{2}{\sigma^2_{CE}}\Bigg( e^{\frac{\sqrt{2}}{\alpha_t\sigma_{CE}}y_t}\text{erfc}\left(\frac{\sigma_t}{\alpha_t\sigma_{CE}} + \frac{y_t}{\sqrt{2}\sigma_t}\right) \\
    + e^{-\frac{\sqrt{2}}{\alpha_t\sigma_{CE}}y_t}\text{erfc}\left(\frac{\sigma_t}{\alpha_t\sigma_{CE}} - \frac{y_t}{\sqrt{2}\sigma_t}\right)\Bigg) \\
    - \frac{4\,e^{-\left(\frac{\sigma_t^2}{\alpha_t^2\sigma_{CE}^2} + \frac{y_t^2}{2 \sigma_t^2}\right)}}{\sqrt{\pi} \alpha_t\sigma_{CE}\sigma_t} \Bigg] \quad \Bigg/ \\
    \Bigg[ e^{\frac{\sqrt{2}}{\alpha_t\sigma_{CE}}y_t}\text{erfc}\left(\frac{\sigma_t}{\alpha_t\sigma_{CE}} + \frac{y_t}{\sqrt{2}\sigma_t}\right) \\
    + e^{-\frac{\sqrt{2}}{\alpha_t\sigma_{CE}}y_t}\text{erfc}\left(\frac{\sigma_t}{\alpha_t\sigma_{CE}} - \frac{y_t}{\sqrt{2}\sigma_t}\right)\Bigg].
\end{multline}

\newpage

\begin{figure*}[t!]
    \centering
    \includegraphics[width=0.72\textwidth]{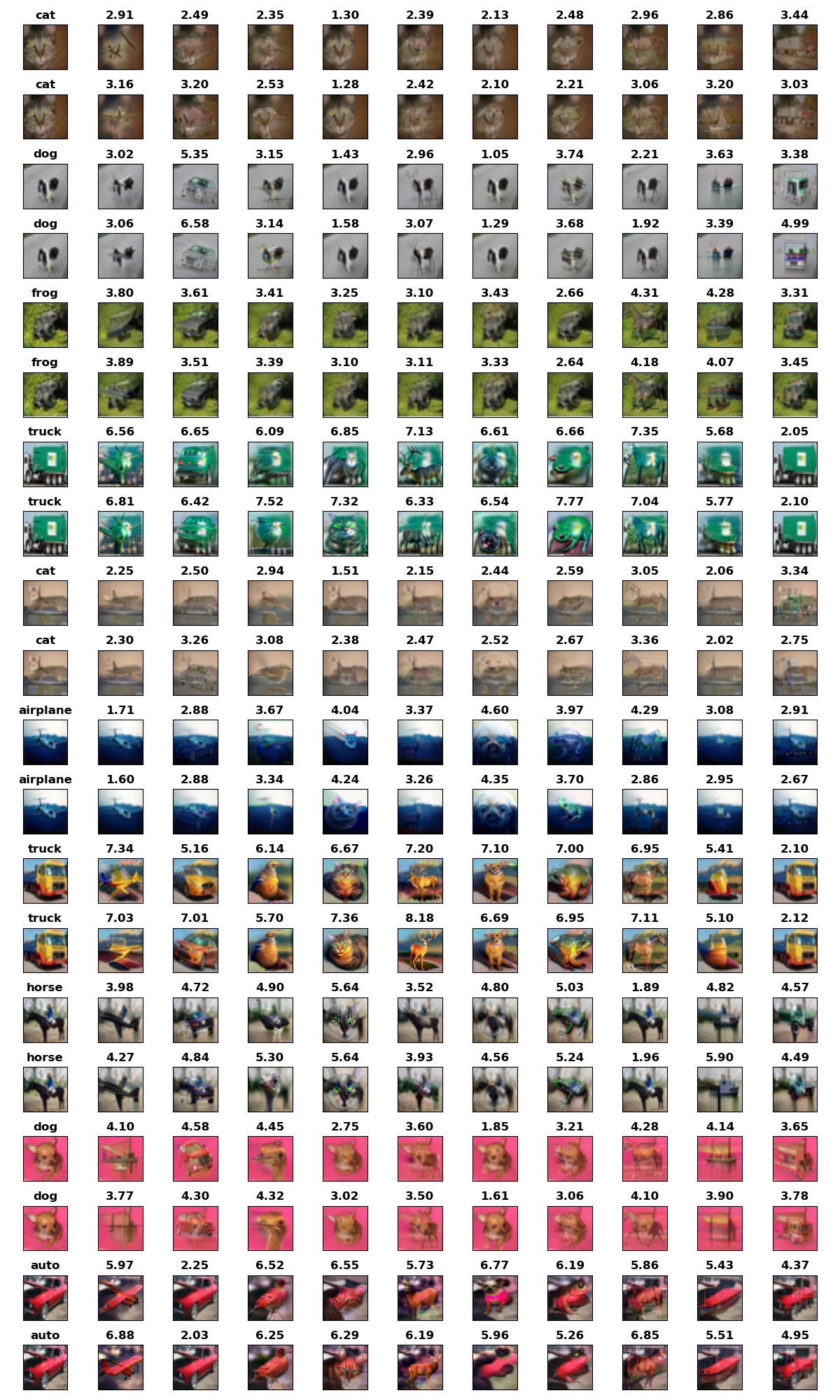}
    \caption{Boltzmann ($w=15$ $\sigma_{CE}=0.2$) CEs from the CIFAR10 training set. Leftmost column: original samples and class label. Other columns: CE and $L^2$ distance to the original sample.}
\end{figure*}

\begin{figure*}[t!]
    \centering
    \includegraphics[width=0.72\textwidth]{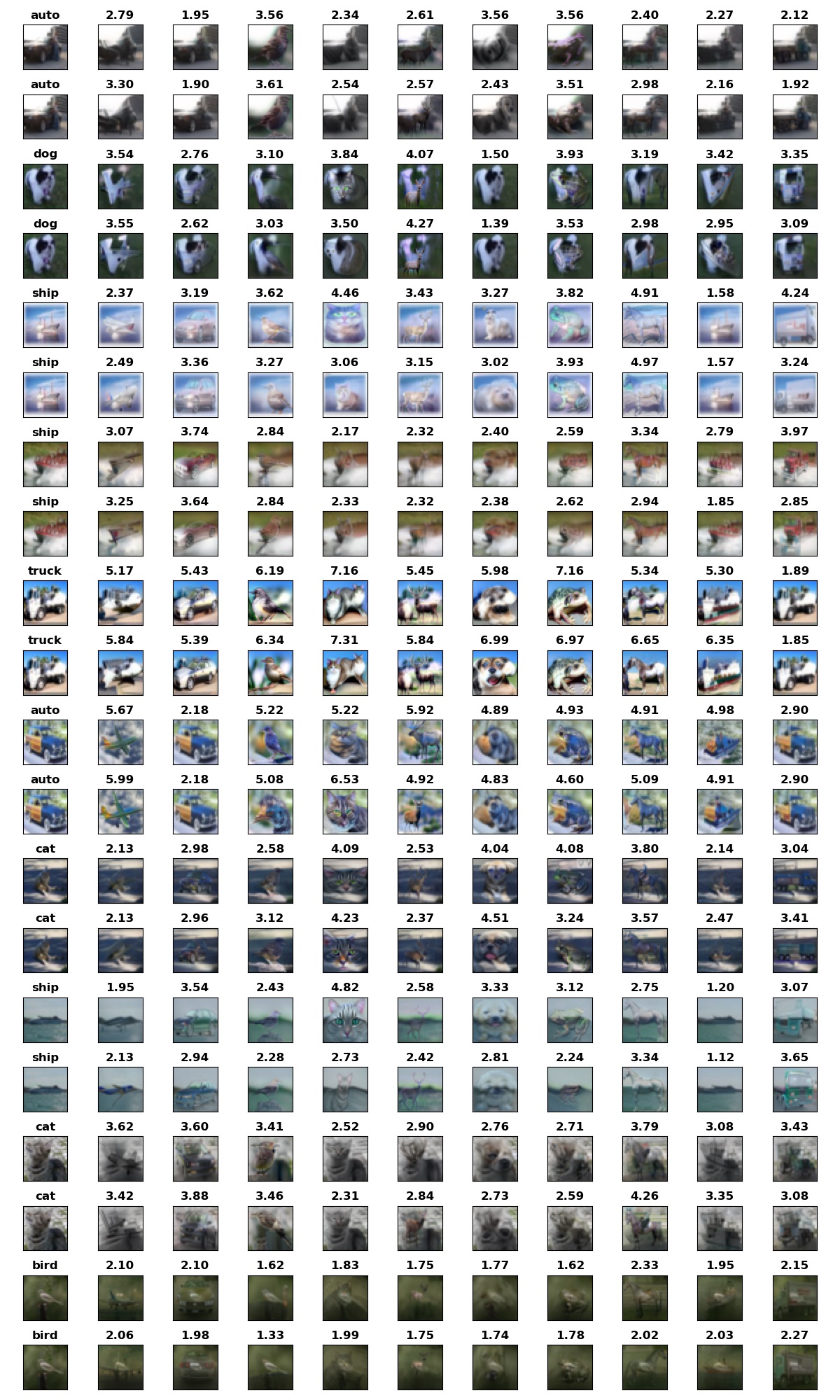}
    \caption{Boltzmann ($w=15$ $\sigma_{CE}=0.2$) CEs from the CIFAR10 training set. Leftmost column: original samples and class label. Other columns: CE and $L^2$ distance to the original sample.}
\end{figure*}

\begin{figure*}[t!]
    \centering
    \includegraphics[width=0.72\textwidth]{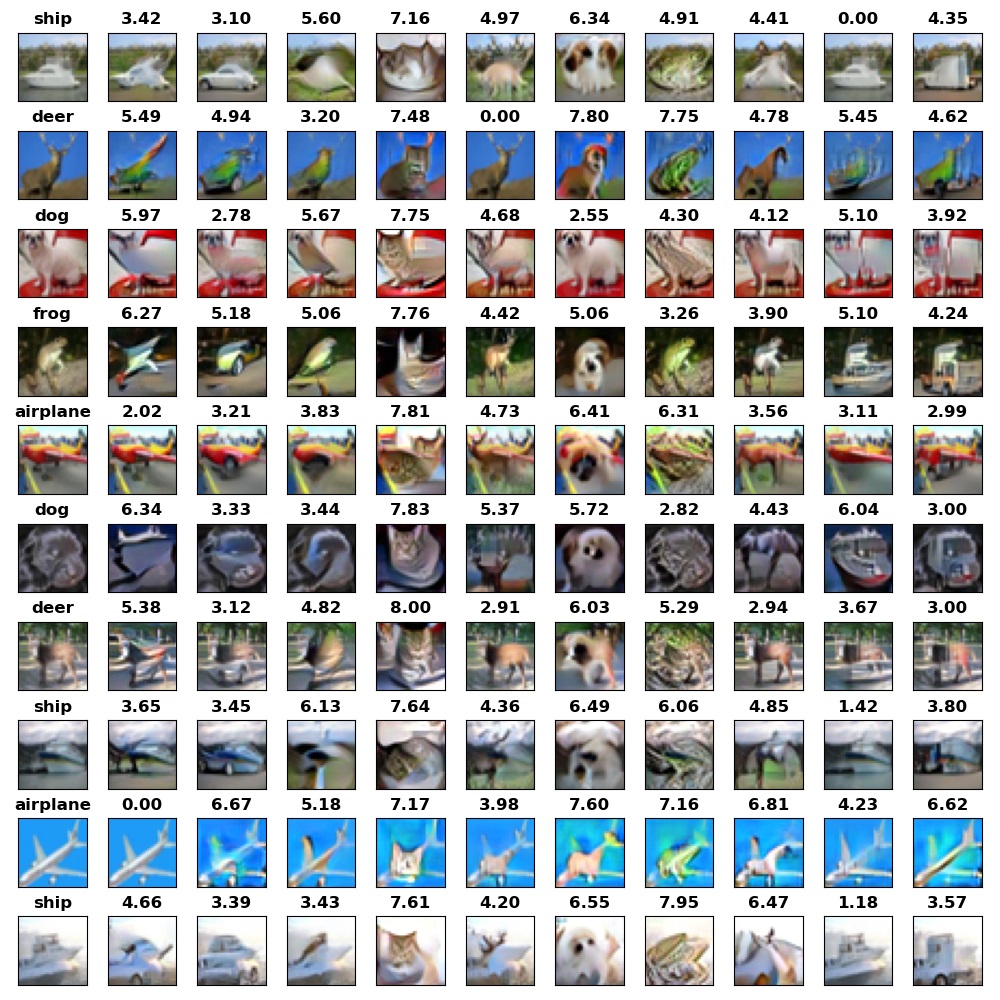}
    \caption{Robust model ($\varepsilon=1$, confidence threshold 0.9) CEs from the CIFAR10 training set. Leftmost column: original samples and class label. Other columns: CE and $L^2$ distance to the original sample.}
\end{figure*}

\begin{figure*}[t!]
    \centering
    \includegraphics[width=0.72\textwidth]{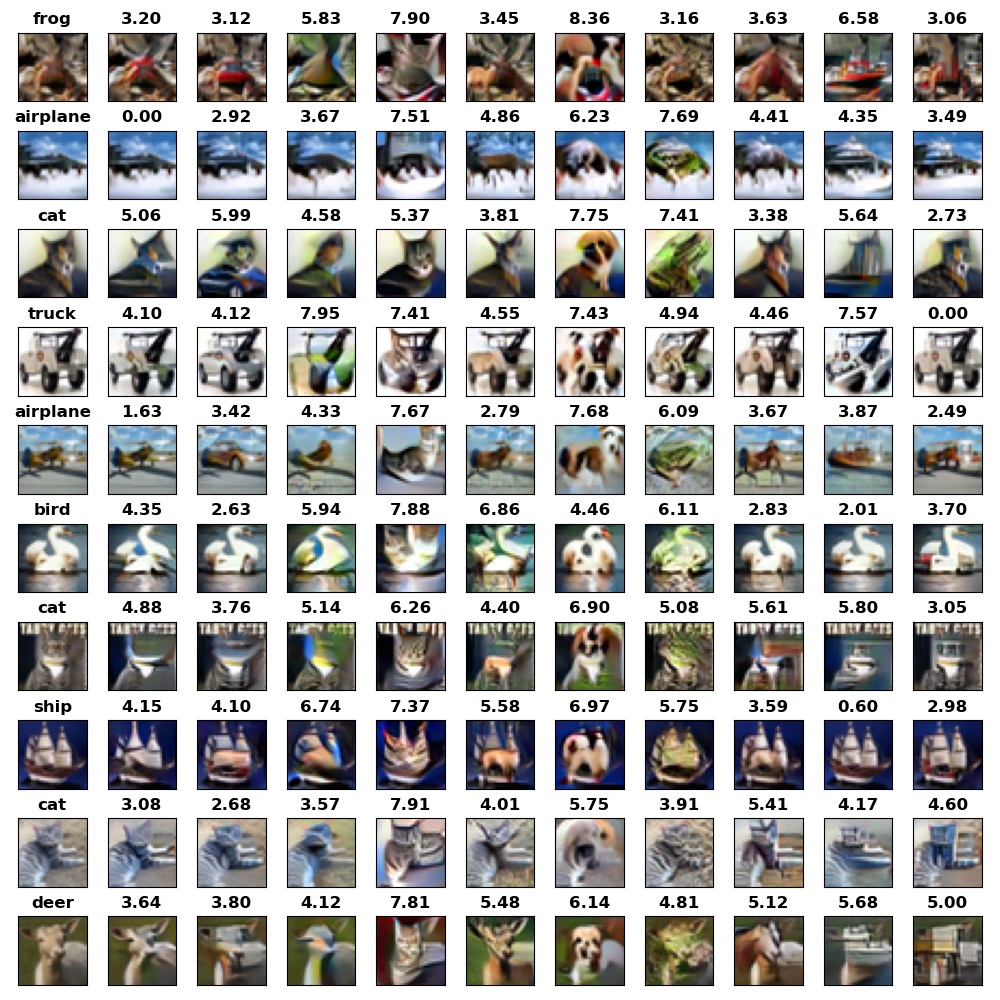}
    \caption{Robust model ($\varepsilon=1$, confidence threshold 0.9) CEs from the CIFAR10 training set. Leftmost column: original samples and class label. Other columns: CE and $L^2$ distance to the original sample.}
\end{figure*}

\begin{figure*}[t!]
    \centering
    \includegraphics[width=0.72\textwidth]{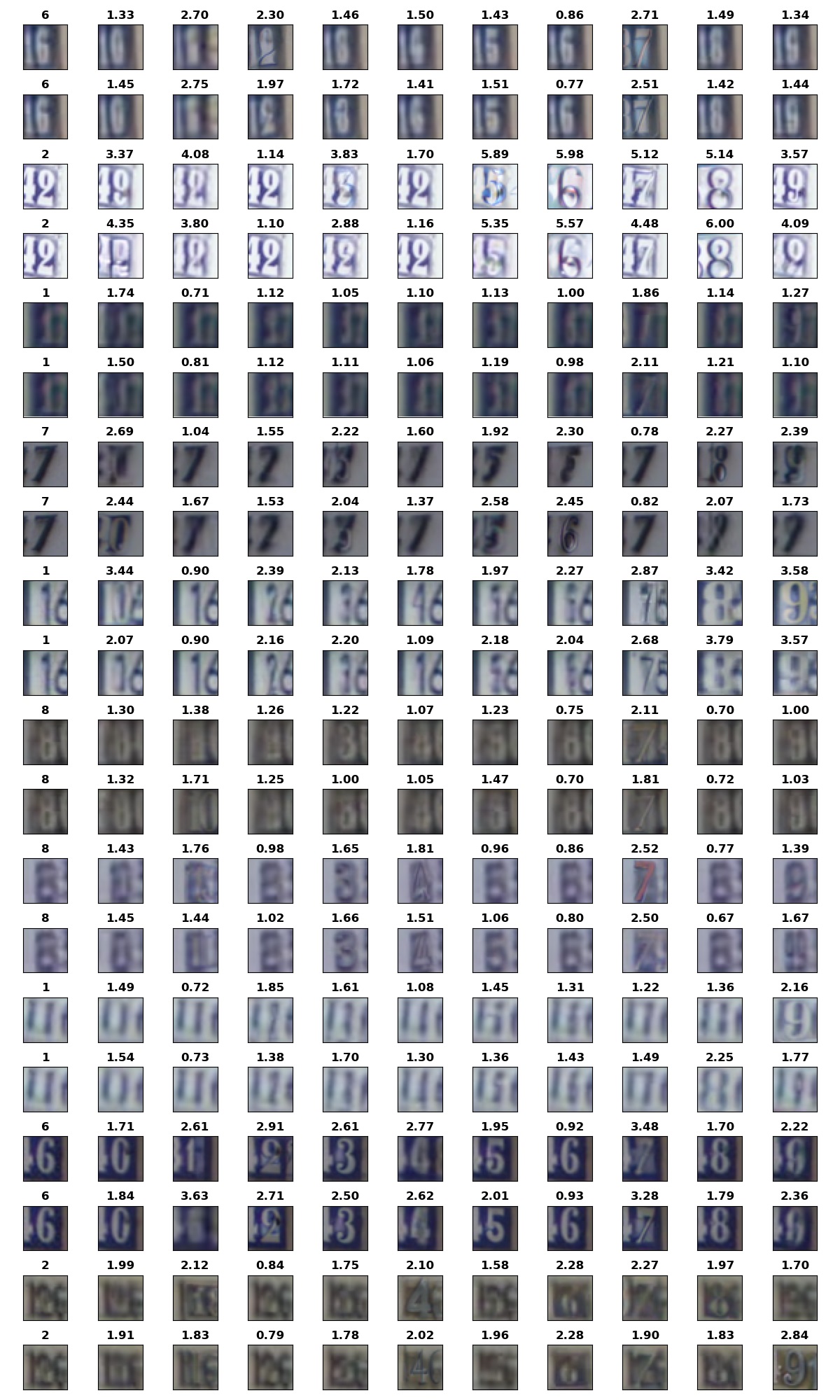}
    \caption{Boltzmann ($w=15$ $\sigma_{CE}=0.2$) CEs from the SVHN training set. Leftmost column: original samples and class label. Other columns: CE and $L^2$ distance to the original sample.}
\end{figure*}

\begin{figure*}[t!]
    \centering
    \includegraphics[width=0.72\textwidth]{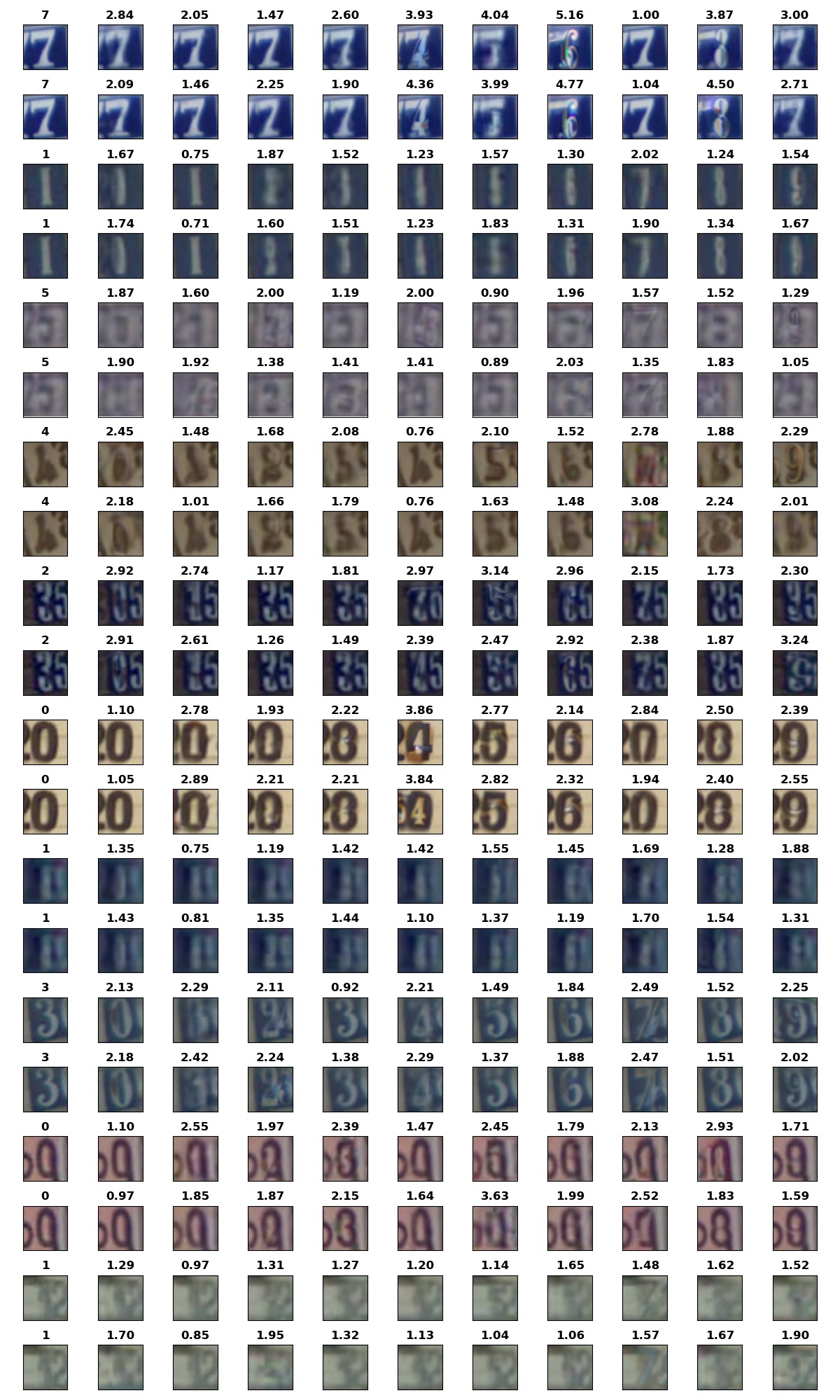}
    \caption{Boltzmann ($w=15$ $\sigma_{CE}=0.2$) CEs from the SVHN training set. Leftmost column: original samples and class label. Other columns: CE and $L^2$ distance to the original sample.}
\end{figure*}

\begin{figure*}[t!]
    \centering
    \includegraphics[width=0.72\textwidth]{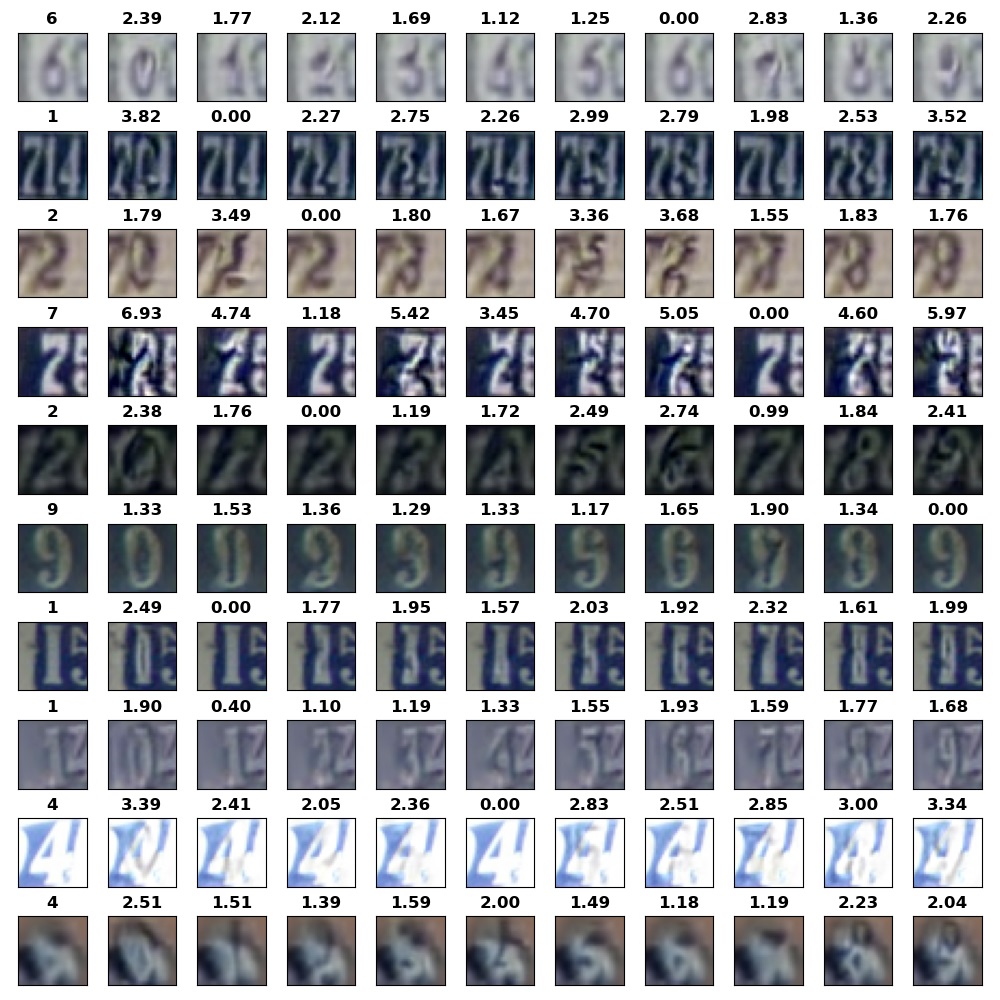}
    \caption{Robust model ($\varepsilon=0.4$, confidence threshold 0.9) CEs from the SVHN training set. Leftmost column: original samples and class label. Other columns: CE and $L^2$ distance to the original sample.}
\end{figure*}

\begin{figure*}[t!]
    \centering
    \includegraphics[width=0.72\textwidth]{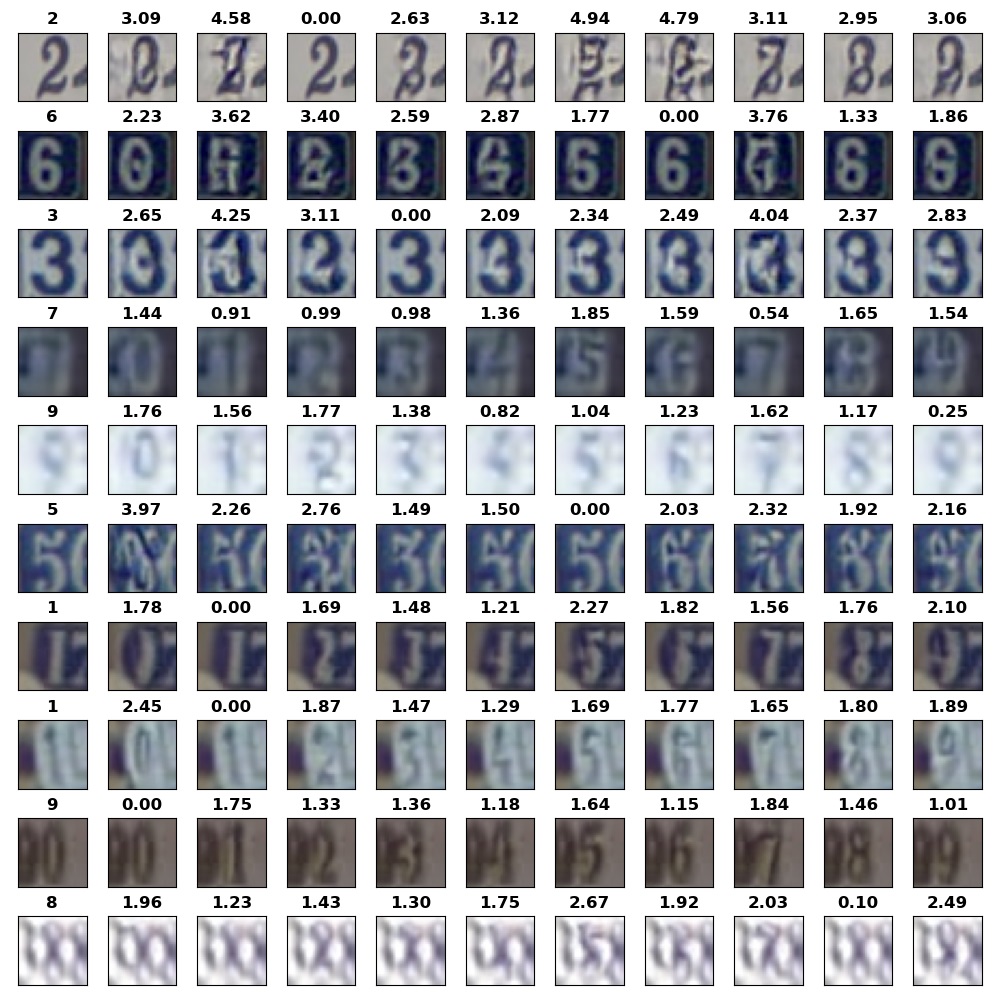}
    \caption{Robust model ($\varepsilon=0.4$, confidence threshold 0.9) CEs from the SVHN training set. Leftmost column: original samples and class label. Other columns: CE and $L^2$ distance to the original sample.}
\end{figure*}

\end{document}